\documentclass[11pt]{article}
\usepackage{acl2012}
\usepackage{times}
\usepackage{latexsym}
\usepackage{amsmath}
\usepackage{multirow}
\usepackage{graphicx}
\usepackage{url}
\DeclareMathOperator*{\argmax}{arg\,max}
\usepackage{ucs}
\usepackage[utf8x]{inputenc}
\usepackage[T1]{fontenc}

\title{How big is big enough?\\Unsupervised word sense disambiguation using a very large corpus}

\author{
Piotr Przybyła \\
  Institute of Computer Science, Polish Academy of Sciences, \\
 Warsaw, Poland, \\
 National Centre for Text Mining, University of Manchester,\\
 Manchester, UK\\
  {\tt piotr.przybyla@manchester.ac.uk}
}

\date{}

\begin{document}
\maketitle
\begin{abstract}
In this paper, the problem of disambiguating a target word for Polish is approached by searching for related words with known meaning. These \textit{relatives} are used to build a training corpus from unannotated text. This technique is improved by proposing new rich sources of replacements that substitute the traditional requirement of monosemy with heuristics based on wordnet relations. The naïve Bayesian classifier has been modified to account for an unknown distribution of senses. A corpus of 600 million web documents (594 billion tokens), gathered by the NEKST search engine allows us to assess the relationship between training set size and disambiguation accuracy. The classifier is evaluated using both a wordnet baseline and a corpus with 17,314 manually annotated occurrences of 54 ambiguous words.
\end{abstract}

\section{Introduction}
The focus of the word sense disambiguation (WSD) task is polysemy, i.e. words having several substantially different meanings. Two common examples are \textit{bank} (riverside or financial institution) and \textit{bass} (fish or musical instrument), but usually the meanings of a word are closely related, e.g. \textit{class} may refer to: (a) a group of students, (b) the period when they meet to study or (c) a room where such meetings occur. Readers deal with this problem by using a word's context and in WSD we aim at doing it automatically.

The most effective solution, called supervised WSD, is to use a large number of sense-annotated occurrences of the target word to build a machine learning model to label test cases. However, this approach suffers from a \textit{knowledge acquisition bottleneck}. The annotation of a separate training corpus for every target word demands a considerable amount of human labour. Therefore, this approach is unusable in applications that require WSD across a wide vocabulary, such as open-domain question answering \cite{Przybya2015}.

The method of \textit{monosemous relatives}, which is the focus of this work, bypasses the bottleneck by gathering occurences of words related to the target word, but free from ambiguity, and treating them as training cases of the respective senses. Human labour is eliminated at the expense of accuracy, as the context of each relative only approximately matches the context of the target word sense.

Monosemous relatives have been employed multiple times (see Section 2), but results remain unsatisfactory. The aim of my study is to explore the limitations of this technique by implementing and evaluating such a tool for Polish. Firstly, the method is expanded by waiving the requirement of monosemy and proposing several new sources of relatives. These previously unexplored sources are based on wordnet data and help gather many training cases from the corpus. Secondly, a well-known problem of uneven yet unknown distribution of word senses is alleviated by modifying a naïve Bayesian classifier. Thanks to this correction, the classifier is no longer biased towards senses that have more training data. Finally, a very large corpus (600 million documents), gathered from the web by a Polish search engine NEKST\footnote{\url{http://www.nekst.pl/}}, is used to build models based on training corpora of different sizes. Those experiments show what amount of data is sufficient for such a task. The proposed solution is compared to baselines that use wordnet structure only, with no training corpora.

This paper is organised as follows. The next section reviews the previous research in the area, focusing on unsupervised WSD using monosemous relatives. Section 3 outlines the proposed solution by describing the new sources of relatives, the employed corpus, the features extracted from context and the modified Bayesian classifier. Section 4 describes the evaluation data and process, while section 5 quotes the results. Section 6 is devoted to discussing the outcomes and section 7 concludes the paper.

\section{Related work}

The problem of WSD has received a lot of attention since the beginning of natural language processing research. WSD is typically expected to improve the results of real-world applications: originally machine translation and recently information retrieval and extraction, especially question answering \cite{Przybya2015}. Like many other areas, WSD has greatly benefited from publicly available test sets and competitions. Two notable corpora are: 1) \textit{SemCor} \cite{Miller1993}, built by labelling a subset of Brown corpus with \textit{Princeton WordNet} synsets and 2) the public evaluations of \textit{Senseval} workshops \cite{snyder-palmer:2004:Senseval-3,mihalcea-chklovski-kilgarriff:2004:Senseval-3}.

There are a variety of approaches to solve the WSD problem, which can be grouped based upon how they use their data -- see reviews \cite{Navigli2009,VidhuBhala2012}. In supervised solutions a large sense-tagged corpus is available for training. This approach has been applied to the the test set used in the current study, resulting in an accuracy value of 91.5\% \cite{Przepiorkowski2012}. Although this technique undoubtedly yields the best results, we would need an immense amount of human labour to build a training corpus of sufficient size for disambiguating all words. This does not seem possible, especially in the case of languages, such as Polish, which receive less attention than English.

In the minimally supervised approach \cite{Yarowsky1995}, a small set of initial training examples, obtained by a heuristic or hand-tagging, is used to label new occurrences. They in turn serve as a training set for next iteration, and so on. This bootstrapping procedure requires very little manual tagging but needs to be carefully implemented to avoid loosing accuracy in further steps. 

Unsupervised methods use no previously labelled examples. Instead an external knowledge source is employed, e.g. a machine-readable dictionary or wordnet. In the simplest unsupervised solution, called the Lesk algorithm \cite{Lesk1986}, meanings of consecutive ambiguous words are selected by finding those senses whose definitions overlap the most.

If lack of definitions make the Lesk algorithm infeasible, we can exploit relations between words. This study focuses on \textit{monosemous relatives}, i.e. words or collocations, selected using wordnet, being related to a disambiguation target, but free of ambiguity. One can easily find occurrences of such relatives in an unannotated text and treat them as training examples for the target ambiguous word. The method has been successfully applied in an English WSD task \cite{Leacock1998}, but still many problems remain. One of them is choice of relatives -- in fact, even synonyms differ in meaning and usage contexts; and they are not available for many words. That is why also hypernyms and hyponyms, especially multi-word expressions containing the target word, are taken into account. Some researchers also include siblings (i.e. words with a common hypernym with the target) and antonyms, but their influence is not always beneficiary \cite{Seo2004}. Other interesting sources of monosemous relatives are parts of definition \cite{Mihalcea1999}, named entities \cite{Mihalcea2000}, indirect hyponyms and hypernyms, and finally meronyms and holonyms \cite{Seo2004}.

The majority of classification techniques are built on an assumption that the training data approximately reflects the true distribution of the target classes. However, that is not the case when using monosemous relatives. The number of their occurrences seldom agrees with the probabilities of corresponding word senses.  Quite often it actually is the opposite: obvious and frequent meanings have very few relatives and vice versa. Some researchers simply copy the \textit{a priori} probabilities from test data  \cite{Leacock1998}, others employ heuristics, but they are easily beaten by statistics taken from a real annotated corpus, even different than test set \cite{Agirre1998}.

Preparing a corpus for finding relatives poses a challenge as well. It should contain a lot of text, as many monosemous words are scarce. Some researchers use snippets retrieved from search engines, i.e. AltaVista \cite{Mihalcea1999} or Google \cite{Agirre1998}. One can also extend a search query by including the context of the disambiguated word \cite{Martinez2006}, but it requires using as many queries as test cases.

Finally, the usage of monosemous relatives has more applications than classical WSD. One can use them to generate topical signatures for concepts \cite{Agirre2001}, automatically build large sense-tagged corpora \cite{Mihalcea2002} and evaluate the quality of wordnet-related semantic resources \cite{Cuadros2006}.

\section{Method}

The algorithm works as follows. First, a set of relatives is obtained for each sense of a target word using the Polish wordnet: \textit{plWordNet} \cite{Maziarz2012}. Some of the replacements may have multiple senses, however usually one of them covers \textit{most} cases. Secondly,  a set of context features is extracted from occurrences of relatives in the NEKST corpus. Finally, the aggregated feature values corresponding to target word senses are used to build a naïve Bayesian classifier adjusted to a situation of unknown \textit{a priori} probabilities.

\subsection{Relatives}
\label{relatives}
\begin{figure}
  \centering
    \includegraphics[width=8cm]{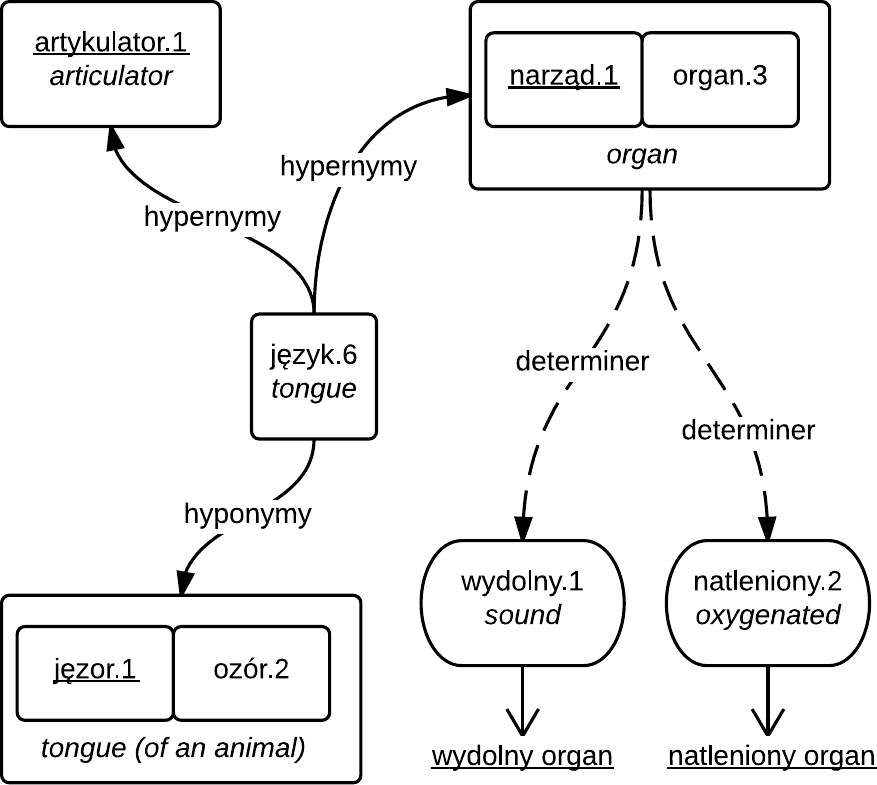}
  \caption{A part of \textit{plWordNet} network used to extract replacements for the word \textit{język} in its 6th sense, meaning tongue (animal or human organ). The resulting replacements are underlined.}
  \label{fig:relatives}  
\end{figure}

In order to obtain training cases from unannotated corpora, we aim to select relatives\footnote{The notion of \textit{relative} covers both individual words and multi-word nominal expressions.} which are semantically similar to a given sense of a target word. An example of this process, concerning the word \textit{język} (tongue) in one of its meanings (human or animal organ) is shown in Figure \ref{fig:relatives}. This study takes into account only synonyms, hypernyms and hyponyms, as other options (siblings, antonyms, higher-order relatives) have previously given unsatisfactory results \cite{Seo2004}. Instead, another problem deserves more attention: how do we select those occurrences of a polysemous relative that correspond to a target word sense? So far, the problem has been circumvented by including only monosemous relatives (\textit{narząd} and \textit{jęzor} in the example), which greatly decreases their availability. Instead, we employ those relatives, whose first meaning is related to the considered sense (\textit{artykulator} in the example). The intuition is that \textit{plWordNet} usually mentions the most frequent meaning as the first.

We also exploit \textit{plWordNet} relations called \textit{determiner}, which links nominals with adjectives that are frequently used to describe them. For example, consider a word \textit{organ} (Eng. organ). An adjective \textit{natleniony} (oxygenated) is a determiner of one of the meanings of \textit{organ}, i.e. part of body, but not the others, i.e. part of an institution. Therefore, the included relatives consist of a polysemous related word, including the target word itself, and a determiner associated with a meaning (\textit{wydolny organ} and \textit{natleniony organ} in the example). This procedure is performed only in the case of relatives that weren't included so far, i.e. with a sense number higher than~1.
 
Finally, we also make use of a well-known principle called \textit{one word per discourse} \cite{Gale1992}, which states that a polysemous word is very unlikely to take different meanings in a single document. In this study, the principle is employed in the following way: if in a single document there appear only relatives corresponding to a single target word sense, then all occurrences of the target word in this document are also treated as training examples for this meaning.

One can easily see that these assumptions are false in many cases, which may  introduce noise and deteriorate a resulting model. Thus, the presented solutions undergo experimental validation using the following sets of relatives:
\begin{enumerate}
\item \textbf{Monosemous children} -- monosemous direct hyponyms.
\item \textbf{Monosemous relatives} -- monosemous synonyms, direct hyponyms and direct hypernyms.
\item \textbf{First relatives} -- words in the first meaning belonging to synsets of synonyms, direct hyponyms or direct hypernyms.
\item \textbf{Word determiners} -- collocations made of two words in any order: the target word and a determiner associated with a given meaning.
\item \textbf{All determiners} -- collocations made of two words in any order: a polysemous relative and a determiner associated with the appropriate meaning.
\item \textbf{Other words} -- occurrences of the target ambiguous word in a document that contains other relatives corresponding to exactly one of the meanings.
\end{enumerate}
Table \ref{tab:relatives} shows how many relatives have been obtained for each category, as well as the number of occurrences in the corpus of 6 million documents (see next section).

\subsection{Corpus}
As some of the relatives may be very rare, it is important to use a training corpus of significant size. In this case, we used 600 million webpages (594 billion tokens) indexed by a Polish search engine NEKST, developed at the Institute of Computer Science, Polish Academy of Sciences. Training sub-corpora were selected with size varying from 19,000 to 60 million documents. Instead of using snippets returned by a search interface, we use whole textual contents (with morphosyntactic annotation) of each document, taken from NEKST distributed infrastructure.

Unfortunately, a lot of text on web pages is not suitable for training a WSD classifier, for example elements of page structure or parts unrecognised by a tagger. Thus, each sentence has to satisfy the following requirements to qualify for training:
\begin{itemize}
\item be at least 150-character long.
\item contain at least five words.
\item contain at least four different parts of speech (including punctuation).
\end{itemize}
These criteria help to filter out most of the web content of unsatisfactory quality.

\subsection{Features}

Context features extracted for classification are very similar to those that have proven successful in supervised WSD, including experiments on the same evaluation set \cite{Modzki2012}:
\begin{itemize}
\item words present at certain positions in a neighbourhood of a target word:
\begin{itemize}
\item lemmas at positions: -2, -1, 1, 2 (denoted by $L_p$),
\item morphosyntactic interpretations (sequences of tags\footnote{Nominal inflection in Slavonic languages, such as Polish, is richer than in English; therefore a sequence of tags is necessary to describe the morphosyntactic interpretation of a single word.}) at positions: -1, 1 (denoted by $I$) and 0 (denoted by $I_0$),
\end{itemize}
\item lemmas present in the sentence (denoted by $L$).
\end{itemize}
Note that the morphosyntactic interpretations are assigned to single words only, therefore in case of multi-word relatives $I_0$ is not available. Also, a gender tag is removed from $I_0$.

\subsection{Classification}
\label{class}

After gathering the values of features from occurrences of relatives, a naïve Bayesian classification model is built. However, as many researchers in the field have noticed \cite{Agirre1998,Leacock1998,Mihalcea1999}, there is a major drawback of using relatives instead of a target word: the number of occurrences of relatives is usually not proportional to the frequency of the target word sense. To bypass this problem, we modify the Bayesian model in this study. The basic classifier chooses the class that maximises the following probability:
\[
P(Y|\mathbf{X})\propto P(Y) \prod_{i=1}^p P(X_i|Y)
\]  
In the case of binary features, which represent the occurrence of a certain word in context, we have $x_i\in \{0,1\}$ and:
\begin{align*}
P(&Y|\mathbf{X}=\mathbf{x})\propto\\
& P(Y)\prod_{i=1}^p\left[ P(X_i=1|Y)^{x_i}P(X_i=0|Y)^{(1-x_i)}\right]
\end{align*}
Which could be rewritten as:
\begin{align*}
P(&Y|\mathbf{X}=\mathbf{x})\propto \\
&\underbrace{P(Y)\prod_{i=1}^p P(X_i=0|Y)}_{A(Y)}\times \underbrace{\prod_{i\,:\,x_i=1}\frac{P(X_i=1|Y)}{P(X_i=0|Y)}}_{B(Y,\mathbf{x})}
\end{align*}
The expression has been formulated as a product of two factors: $A(Y)$, independent from observed features and corresponding to empty word context, and $B(Y,\mathbf{x})$ that depends on observed context. To weaken the influence of improper distribution of training cases, we omit $A(Y)$, so that when no context features are observed, every word sense is considered equally probable.

Thus, for the given context features $\mathbf{x}$, sense $y^*$ is selected when:
\[
y^*=\argmax_y \prod_{i\,:\,x_i=1}\frac{P(X_i=1|Y=y)}{P(X_i=0|Y=y)}
\] 
The table with final results (\ref{tab:baselines}) contains accuracies of both original and modified versions of the classifier.

\section{Evaluation}

For experiments and evaluation a sub-corpus of the National Corpus of Polish, NCP \cite{Przepiorkowski2012} was employed. The manually annotated sub-corpus contains sense labels of 106 different words: 50 nouns, 48 verbs and 8 adjectives. As verbs have much poorer connectivity in \textit{plWordNet}, they have been ignored within this study.

The senses used for annotation are coarse-grained -- with one sense covering a range of related usages. Each word has between two and four senses. To employ the method described in previous section, the NCP senses have been manually mapped to fine-grained \textit{plWordNet} synsets. As NCP senses have been created independently from wordnet senses, a substantial part of the latter remain uncovered by the former. However, we only found four cases where an NCP sense has no counterpart in wordnet; those words are excluded from the test set.

In total, the test set includes 17,314 occurrences of 54 ambiguous words, having two to four coarse-grained meanings. Table \ref{tab:words} contains a detailed summary.

The algorithm works using \textit{plWordNet} synsets and its output is mapped to NCP meaning to measure its quality. Accuracy measures what percentage of the programme's guesses agree with manually assigned senses. To assess the general performance of a particular configuration, the accuracy has been averaged over all target words.

To properly judge the results, we need to start with the baselines. Without knowing the distribution of senses, three basic possibilities seem reasonable: we can either 1) select a meaning randomly, 2) base on the sense numbering in NCP or 3) use \textit{plWordNet} in the same way. To have a better comparison with ontology-based methods, the results also include a word similarity baseline configuration, which selects the sense with the strongest similarity to any of the words in context (sentence). For that purpose the Leacock\&Chodorow similarity measure (implemented using all relations between synsets in \textit{plWordNet}) is employed, as it has been previously used in WSD \cite{Leacock1998a} and also correlates well with human judgement of similarity \cite{Budanitsky2006}. The baseline results, shown in Table \ref{tab:baselines}, support the claim of intentional sense ordering in \textit{plWordNet}.

\section{Results}

\begin{table}
\begin{center}
\begin{tabular}{r c c c}
\hline\hline \bf Feature set & \bf  Mean accuracy \\ 
\hline
$I_0$ &  69.32\% \\
$I$ & 69.42\% \\
$L$ &  75.69\% \\
$L_p$ & 74.16\% \\
$L$, $L_p$ &  77.45\% \\
$L$, $L_p$, $I$  &\bf 77.96\% \\
$L$, $L_p$, $I$, $I_0$  & 77.66\% \\
\hline \hline
\end{tabular}
\end{center}
\caption{Mean accuracy of the disambiguation algorithm with respect to the features involved in the classification ($I_0$ -- interpretation of a disambiguated word, $I$ -- interpretations of neighbouring words, $L_p$ -- lemmas of neighbouring words, $L$ -- lemmas present in the whole sentence).}
\label{tab:features}
\end{table}

The goal of the first experiment was to select an optimal feature set for this task. Several models with a common basic configuration, i.e. using all possible relatives and 6 million documents, have been built with different feature sets and evaluated. The results are shown in Table \ref{tab:features}. As we can see, the lexical features give us more predictive power than morphological interpretations. The best solution, incorporated into the basic configuration for further experiments, includes all features except these based on the interpretation of the word in focus.

\begin{table*}
\begin{center}
\begin{tabular}{l c c c}
\hline\hline \bf Type of replacements & \bf Replacements & \bf Occurrences in corpus & \bf Mean accuracy \\ 
\hline
Monosemous children & 25.48 & 383,498 & 63.31\% \\
+ Monosemous relatives & 30.09 & 769,947 & 70.86\% \\
+ First relatives & 44.89 & 2,295,686 & 77.35\% \\
+ Word determiners &  51.26 & 2,296,514 & 77.72\% \\
+ All determiners &  102.04 & 2,309,640 & 77.64\% \\
+ Other words &  103.04 & 2,473,255 & \bf 77.96\% \\
\hline \hline
\end{tabular}
\end{center}
\caption{Strategies for generating replacements, each built by adding new elements to the previous step, with the resulting number of replacements (average per word), their occurrences in the corpus (total) and the mean accuracy of disambiguation.}
\label{tab:relatives}
\end{table*}

Secondly, it is necessary to evaluate different types of replacements, outlined in section \ref{relatives}. Table \ref{tab:relatives} contains the average number of possible replacements per target word, the number of occurrences in the six-million corpus and the average classification accuracy. As we can see, the number of replacements rises after adding subsequent sources, but the largest increase is caused by including polysemous relatives with determiners. On the other hand, these compound relatives rarely appear in the corpus (13,126 occurences), whereas employing polysemous words in the first sense results in 1,525,739 new training cases and a substantial growth of accuracy. What is more, although the profits from these sources of relatives differ, none of them decreases the accuracy.

\begin{figure*}
  \centering
    \includegraphics[width=15cm]{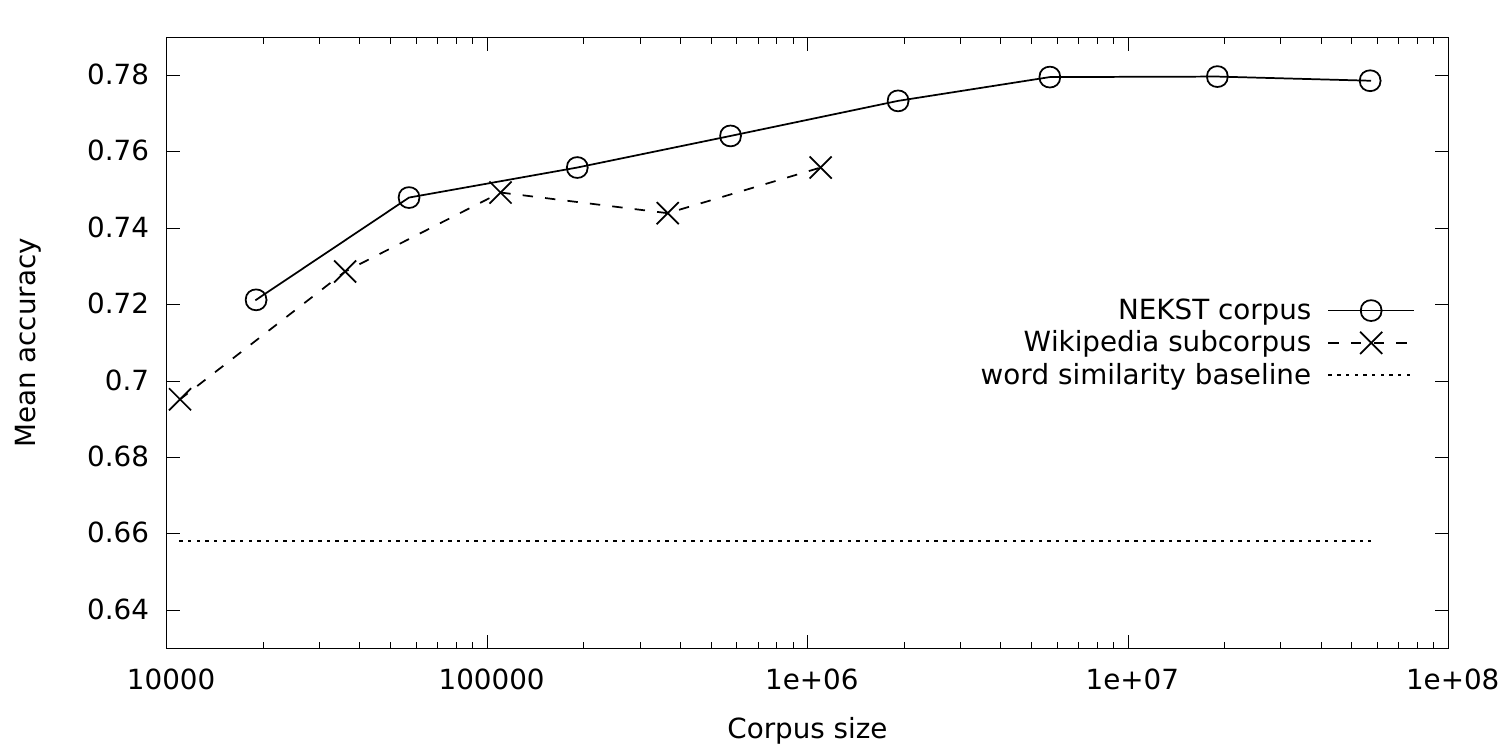}
  \caption{Mean disambiguation accuracy of models built using corpora of different sizes, created by random selection from 600 million web documents from NEKST search engine: unrestricted or only from the Polish Wikipedia. The best of the baselines, which uses wordnet-based word similarity is also shown. }
  \label{fig:plot}  
\end{figure*}

The availability of the corpus containing 600 million documents helps to answer the question of sufficient corpus size for such task. Figure \ref{fig:plot} shows mean classification accuracy for models built using different training corpora, which have been created by randomly selecting a subset of the original document set. The considered sizes are between 19,000 and 60,000,000\footnote{The whole corpus hasn't been tested because of insufficient computational resources.}. Additionally, a different type of corpora has been created by using only documents from the Polish Wikipedia (sizes 11,000 -- 1,098,000). We see that after a certain point adding in new data does not improve the accuracy. Surprisingly, the subcorpora of Wikipedia yield worse results than those of unlimited origin.

\begin{table}
\footnotesize
\begin{center}
\begin{tabular}{r c c c}
\hline\hline \bf Word & \bf  Meanings & \bf Test cases &\bf Accuracy \\ 
\hline
sztuka & 3 & 232 & 73.27\% \\
sprawa & 2 & 1499 & 64.17\% \\
raz & 3 & 1358 & 91.60\% \\
państwo & 2 & 714 & 85.29\% \\
strona & 3 & 760 & 53.94\% \\
rada & 3 & 561 & 92.15\% \\
stan & 4 & 480 & 83.95\% \\
rząd & 2 & 570 & 58.94\% \\
punkt & 4 & 268 & 56.71\% \\
izba & 2 & 177 & 53.67\% \\
uwaga & 2 & 386 & 82.38\% \\
program & 3 & 518 & 70.65\% \\
ważny & 2 & 488 & 97.33\% \\
akcja & 2 & 368 & 73.09\% \\
zasada & 2 & 350 & 72.00\% \\
działanie & 2 & 265 & 98.11\% \\
mina & 2 & 41 & 90.24\% \\
okres & 2 & 375 & 99.20\% \\
rynek & 2 & 266 & 92.10\% \\
podstawa & 2 & 279 & 78.13\% \\
zakład & 2 & 326 & 80.36\% \\
członek & 3 & 367 & 92.91\% \\
ziemia & 3 & 312 & 65.70\% \\
prosty & 2 & 516 & 94.96\% \\
piłka & 2 & 78 & 100\% \\
bliski & 2 & 279 & 69.17\% \\
kultura & 2 & 276 & 99.63\% \\
stanowisko & 2 & 312 & 68.58\% \\
powód & 2 & 322 & 97.82\% \\
góra & 4 & 243 & 58.02\% \\
wiek & 2 & 324 & 79.93\% \\
forma & 3 & 255 & 64.70\% \\
język & 2 & 258 & 91.08\% \\
wolny & 3 & 168 & 52.97\% \\
wysokość & 2 & 205 & 99.02\% \\
letni & 2 & 50 & 82.00\% \\
klasa & 4 & 197 & 34.01\% \\
zawód & 2 & 86 & 84.88\% \\
stosunek & 3 & 186 & 60.75\% \\
rola & 3 & 249 & 98.39\% \\
skład & 2 & 135 & 85.18\% \\
pole & 3 & 119 & 68.90\% \\
stopień & 3 & 207 & 74.39\% \\
wpływ & 2 & 228 & 33.77\% \\
pokój & 2 & 253 & 43.47\% \\
oddział & 2 & 197 & 83.24\% \\
drogi & 2 & 130 & 83.84\% \\
ciało & 3 & 256 & 76.17\% \\
kolej & 2 & 125 & 77.60\% \\
światło & 2 & 166 & 97.59\% \\
pozycja & 3 & 105 & 76.19\% \\
pismo & 2 & 173 & 79.19\% \\
pozostały & 3 & 131 & 97.70\% \\
doświadczenie & 2 & 125 & 90.40\% \\
\hline
\bf average & 2.44 & 320.63 & 77.96\% \\ 
\hline \hline
\end{tabular}
\end{center}
\caption{Polysemous words used for evaluation with their number of meanings, test cases and obtained disambiguation accuracy in basic configuration.}
\label{tab:words}
\end{table}

Table \ref{tab:words} shows the accuracy of disambiguation in the configuration outlined above with respect to the target word. The easiest words have meanings corresponding to distinct physical objects, e.g. in Polish \textit{piłka} (100\% accuracy) may mean \textit{a ball} or \textit{a small saw}. The hardest cases are those with many abstract and fuzzy meanings, e.g. in Polish \textit{klasa} has four meanings related to English \textit{class}: (1) \textit{group of similar objects}, (2) \textit{level of quality}, (3) \textit{group of pupils} or (4) \textit{classroom}. The meaning (1) could be hard to distinguish from (2) even for a human, whereas (3) and (4) may appear in very similar contexts.

\begin{table}
\begin{center}
\begin{tabular}{r c}
\hline\hline \bf Configuration & \bf  Mean accuracy \\ 
\hline
random baseline &  43.21\% \\
first NCP sense baseline& 58.36\% \\
first \textit{plWordNet} sense baseline &  64.73\% \\
word similarity baseline & 65.81\%\\
\hline
classifier (Bayesian) & 76.89\%\\
classifier (modified) & 77.95\%\\
\hline \hline
\end{tabular}
\end{center}
\caption{Accuracy of four baseline configurations (selecting senses randomly, basing on sense order in the National Corpus of Polish or the Polish wordnet, and choosing the sense which is the most similar to context according to Leacock\&Chodorow measure) and two versions of the classifier proposed in this work (based on the traditional naïve Bayesian model or modified as in section \ref{class}).}
\label{tab:baselines}
\end{table}

Finally, Table \ref{tab:baselines} contains the mean accuracy of the basic configuration of the classifier described in this work (with and without modifications to Bayesian model). It is compared to the four previously mentioned baselines.

\section{Discussion}

Although many different configurations have been tested in this study, all of them remain below the accuracy level of 80\%, approximately equal to average share of dominating senses in this dataset. This is obviously unsatisfactory and demands explanation.

First of all, the new sources of replacements proposed in this work indeed seem to improve the models from 70.86\% (only traditional monosemous relatives) to 77.96\% (all proposed relatives). The biggest gain is obtained by including the polysemous relatives taking into account only their first meaning. This technique relies on two assumptions: a strong domination of one of the senses and that sense being listed first in \textit{plWordNet}. While the former is almost always true, if the second assumption is false then the created model are adversely affected. In the case of two target words the senses, the first sense in each case (\textit{stopień} as a musical concept and \textit{forma} as a synonym of polynomial) was so peculiar that they were unknown to the author of this study and couldn't be assigned to any of the coarse-grained NCP senses. Clearly, not only the method of unsupervised WSD using relatives, but also other solutions related to polysemy would definitely benefit from a reliable ordering of senses in wordnets, especially as increasingly uncommon senses are added to them with time. It is however not clear how such knowledge could be obtained without solving the WSD task first. What is more, sense distributions obviously change with genre, time, author, etc. 

When it comes to feature selection, the most unexpected phenomenon observed in this study is low usefulness of the interpretation-based features. According to Table \ref{tab:features}, adding interpretations of neighbouring words ($I$) yields very little improvement, while this type of information regarding replacements ($I_0$) even lowers the accuracy. This result could be attributed to two factors. Firstly, more developed replacement generation results in more occurrences, but also causes their tags to differ from the target word by gender or number. They may even not be available at all (in the case of multi-word replacements). The second reason is a difference in language: while in English a word interpretation is represented as one of several dozen part of speech identifiers, in Slavonic languages, such as Polish, we need to specify the values of several tags for each word, leading to thousands of possible interpretations. Obviously, the features based on these tags are very sparse. Finally, the morphosyntactic annotation was performed automatically, which may lead to errors, especially in the case of noisy web text.

One of the purposes of this study was to check the necessary amount of training data for such a solution by employing a very large collection from the NEKST search engine. The need for large corpora is obvious when using only monosemous relatives -- those usually rare words should appear in many contexts. However, according to the results shown in Figure \ref{fig:plot}, the strategy for generating relatives presented in this paper reaches optimum performance for a reasonable amount of texts -- 6 million documents is enough. However, one should keep in mind that this statement remains true assuming a constant evaluation environment; expanding a test set (currently containing 17,314 occurrences) may help to see differences between apparently equivalent models and raise the need for bigger corpora.

\section{Conclusions}

In this paper the limitations and improvements of unsupervised word sense disambiguation have been investigated. The main problem -- insufficient number and quality of replacements has been tackled by adding new rich sources of replacements. The quality of the models has indeed improved, especially thanks to replacements based on sense ordering in \textit{plWordNet}. To deal with the problem of unknown sense distribution, the Bayesian classifier has been modified, removing the bias towards frequent labels in the training data. Finally, the experiments with very large corpus have shown the sufficient amount of training data for this task, which is only 6 million documents.  

\section*{Acknowledgements}
This study was conducted at Institute of Computer Science, Polish Academy of Sciences, and supported by a research fellowship within "Information technologies: research and their interdisciplinary applications" agreement number POKL.04.01.01-00-051/10-00. The author would like to thank Dariusz Czerski and NEKST team for providing access to the search engine index and helpful discussions and Matthew Shardlow for comments that greatly improved the manuscript.

\bibliographystyle{acl2012}
\bibliography{biblio}

\begin{thebibliography}{}

\bibitem[\protect\citename{Agirre and Martinez}2004]{Agirre1998}
Eneko Agirre and David Martinez.
\newblock 2004.
\newblock {Unsupervised WSD based on automatically retrieved examples: The
  importance of bias}.
\newblock In {\em Proceedings of the Conference on Empirical Methods in Natural
  Language Processing (EMNLP)}, pages 25--33. Association for Computational
  Linguistics.

\bibitem[\protect\citename{Agirre \bgroup et al.\egroup }2001]{Agirre2001}
Eneko Agirre, Olatz Ansa, Eduard Hovy, and David Martinez.
\newblock 2001.
\newblock {Enriching WordNet concepts with topic signatures}.
\newblock In {\em Proceedings of the NAACL workshop on WordNet and Other
  lexical Resources: Applications, Extensions and Customizations}. Association
  for Computational Linguistics.

\bibitem[\protect\citename{Budanitsky and Hirst}2006]{Budanitsky2006}
Alexander Budanitsky and Graeme Hirst.
\newblock 2006.
\newblock {Evaluating WordNet-based Measures of Lexical Semantic Relatedness}.
\newblock {\em Computational Linguistics}, 32(1):13--47.

\bibitem[\protect\citename{Cuadros and Rigau}2006]{Cuadros2006}
Montse Cuadros and German Rigau.
\newblock 2006.
\newblock {Quality assessment of large scale knowledge resources}.
\newblock In {\em Proceedings of the 2006 Conference on Empirical Methods in
  Natural Language Processing}, pages 534--541. Association for Computational
  Linguistics.

\bibitem[\protect\citename{Gale \bgroup et al.\egroup }1992]{Gale1992}
W.~A. Gale, K.~W. Church, and D.~Yarowsky.
\newblock 1992.
\newblock {One sense per discourse}.
\newblock {\em Proceedings of the DARPA Speech and Natural LanguageWorkshop
  (Harriman, NY)}, pages 233--237.

\bibitem[\protect\citename{Leacock and Chodorow}1998]{Leacock1998a}
Claudia Leacock and Martin Chodorow.
\newblock 1998.
\newblock {Combining Local Context and WordNet Similarity for Word Sense
  Identification}.
\newblock In {\em WordNet: An electronic lexical database.}, pages 265--283.

\bibitem[\protect\citename{Leacock \bgroup et al.\egroup }1998]{Leacock1998}
Claudia Leacock, George~A. Miller, and Martin Chodorow.
\newblock 1998.
\newblock {Using corpus statistics and WordNet relations for sense
  identification}.
\newblock {\em Computational Linguistics}, 24(1):147--165.

\bibitem[\protect\citename{Lesk}1986]{Lesk1986}
Michael Lesk.
\newblock 1986.
\newblock {Automatic sense disambiguation using machine readable dictionaries}.
\newblock In {\em Proceedings of the 5th annual international conference on
  Systems documentation - SIGDOC '86}, pages 24--26. ACM Press.

\bibitem[\protect\citename{Martinez \bgroup et al.\egroup }2006]{Martinez2006}
David Martinez, Eneko Agirre, and Xinglong Wang.
\newblock 2006.
\newblock {Word Relatives in Context for Word Sense Disambiguation}.
\newblock In {\em Proceedings of the 2006 Australasian Language Technology
  Workshop (ALTW2006)}, pages 42--50.

\bibitem[\protect\citename{Maziarz \bgroup et al.\egroup }2012]{Maziarz2012}
Marek Maziarz, Maciej Piasecki, and Stanisław Szpakowicz.
\newblock 2012.
\newblock {Approaching plWordNet 2.0}.
\newblock In {\em Proceedings of the 6th Global Wordnet Conference}. The Global
  WordNet Association.

\bibitem[\protect\citename{Mihalcea and Moldovan}1999]{Mihalcea1999}
Rada Mihalcea and Dan~I. Moldovan.
\newblock 1999.
\newblock {An Automatic Method for Generating Sense Tagged Corpora}.
\newblock In {\em Proceedings of the sixteenth national conference on
  artificial intelligence and the eleventh Innovative applications of
  artificial intelligence conference}.

\bibitem[\protect\citename{Mihalcea and Moldovan}2000]{Mihalcea2000}
Rada Mihalcea and Dan~I. Moldovan.
\newblock 2000.
\newblock {An Iterative Approach to Word Sense Disambiguation}.
\newblock In {\em Proceedings of the Thirteenth International Florida
  Artificial Intelligence Research Society Conference}, pages 219--223. AAAI
  Press.

\bibitem[\protect\citename{Mihalcea \bgroup et al.\egroup
  }2004]{mihalcea-chklovski-kilgarriff:2004:Senseval-3}
Rada Mihalcea, Timothy Chklovski, and Adam Kilgarriff.
\newblock 2004.
\newblock {The Senseval-3 English lexical sample task}.
\newblock In {\em Senseval-3: Third International Workshop on the Evaluation of
  Systems for the Semantic Analysis of Text}, pages 25--28. Association for
  Computational Linguistics.

\bibitem[\protect\citename{Mihalcea}2002]{Mihalcea2002}
Rada Mihalcea.
\newblock 2002.
\newblock {Bootstrapping Large Sense Tagged Corpora.}
\newblock In {\em Proceedings of the Third International Conference on Language
  Resources and Evaluation (LREC-2002)}. European Language Resources
  Association (ELRA).

\bibitem[\protect\citename{Miller \bgroup et al.\egroup }1993]{Miller1993}
George~A. Miller, Claudia Leacock, Randee Tengi, and Ross~T. Bunker.
\newblock 1993.
\newblock {A Semantic Concordance}.
\newblock In {\em Proceedings of the Workshop on Human Language Technology -
  HLT '93}, pages 303--308. Association for Computational Linguistics.

\bibitem[\protect\citename{Młodzki \bgroup et al.\egroup }2012]{Modzki2012}
Rafał Młodzki, Mateusz Kope\'{c}, and Adam Przepi\'{o}rkowski.
\newblock 2012.
\newblock {Word Sense Disambiguation in the National Corpus Of Polish}.
\newblock {\em Prace Filologiczne}, LXIII:155--166.

\bibitem[\protect\citename{Navigli}2009]{Navigli2009}
Roberto Navigli.
\newblock 2009.
\newblock {Word sense disambiguation: A survey}.
\newblock {\em ACM Computing Surveys (CSUR)}, 41(2):10.

\bibitem[\protect\citename{Przepi\'{o}rkowski \bgroup et al.\egroup
  }2012]{Przepiorkowski2012}
Adam Przepi\'{o}rkowski, Mirosław Bańko, Rafał~L. G\'{o}rski, and Barbara
  Lewandowska-Tomaszczyk.
\newblock 2012.
\newblock {\em {Narodowy Korpus Języka Polskiego}}.
\newblock Wydawnictwo Naukowe PWN, Warszawa.

\bibitem[\protect\citename{Przybyła}2015]{Przybya2015}
Piotr Przybyła.
\newblock 2015.
\newblock {Gathering Knowledge for Question Answering Beyond Named Entities}.
\newblock In Chris Biemann, Siegfried Handschuh, Andr\'{e} Freitas, Farid
  Meziane, and Elisabeth M\'{e}tais, editors, {\em Proceedings of the 20th
  International Conference on Applications of Natural Language to Information
  Systems (NLDB 2015)}, pages 412--417, Passau, Germany. Springer-Verlag.

\bibitem[\protect\citename{Seo \bgroup et al.\egroup }2004]{Seo2004}
Hee-Cheol Seo, Hoojung Chung, Hae-Chang Rim, Sung~Hyon Myaeng, and Soo-Hong
  Kim.
\newblock 2004.
\newblock {Unsupervised word sense disambiguation using WordNet relatives}.
\newblock {\em Computer Speech \& Language}, 18(3):253--273.

\bibitem[\protect\citename{Snyder and
  Palmer}2004]{snyder-palmer:2004:Senseval-3}
Benjamin Snyder and Martha Palmer.
\newblock 2004.
\newblock {The English all-words task}.
\newblock In {\em Senseval-3: Third International Workshop on the Evaluation of
  Systems for the Semantic Analysis of Text}, pages 41--43. Association for
  Computational Linguistics.

\bibitem[\protect\citename{{Vidhu Bhala} and Abirami}2012]{VidhuBhala2012}
R.~V. {Vidhu Bhala} and S.~Abirami.
\newblock 2012.
\newblock {Trends in word sense disambiguation}.
\newblock {\em Artificial Intelligence Review}, 42(2):159--171.

\bibitem[\protect\citename{Yarowsky}1995]{Yarowsky1995}
David Yarowsky.
\newblock 1995.
\newblock {Unsupervised word sense disambiguation rivaling supervised methods}.
\newblock In {\em Proceedings of the 33rd annual meeting on Association for
  Computational Linguistics}, pages 189--196. Association for Computational
  Linguistics.

\end{thebibliography}

\end{document}